\documentclass[10pt,twocolumn,letterpaper]{article}

\usepackage{cvpr}              

\usepackage{graphicx}
\usepackage{amsmath}
\usepackage{amssymb}
\usepackage{booktabs}
\usepackage{overpic}

\usepackage{amsthm}
\newtheorem{theorem}{Theorem}

\usepackage[pagebackref,breaklinks,colorlinks]{hyperref}

\usepackage[capitalize]{cleveref}
\crefname{section}{Sec.}{Secs.}
\Crefname{section}{Section}{Sections}
\Crefname{table}{Table}{Tables}
\crefname{table}{Tab.}{Tabs.}

\def\cvprPaperID{9110} 
\def\confName{CVPR}
\def\confYear{2022}

\newcommand{\R}[0]{\mathbb{R}}
\newcommand{\C}[0]{\mathbb{C}}
\newcommand{\diff}{\mathrm{d}}
\newcommand{\divg}{\mathrm{div}}
\newcommand{\I}{\mathbf{g}}

\DeclareMathOperator*{\argmin}{arg\,min}

\usepackage{xcolor}
\definecolor{bgcol}{RGB}{190,220,255}

\newcommand{\mypara}[1]{\vspace{-3mm}\paragraph*{#1}}

\begin{document}


\title{Deep Orientation-Aware Functional Maps:\\ Tackling Symmetry Issues in Shape Matching}

\author{Nicolas Donati\\
LIX, \'Ecole Polytechnique\\
{\tt\small nicolas.donati@polytechnique.edu}
\and
Etienne Corman\\
CNRS, Inria\\
{\tt\small etienne.corman@cnrs.fr}
\and
Maks Ovsjanikov\\
LIX, \'Ecole Polytechnique\\
{\tt\small maks@lix.polytechnique.fr}
}

\maketitle

\begin{abstract}
    State-of-the-art fully intrinsic networks for non-rigid shape matching often struggle to disambiguate the symmetries of the shapes leading to unstable correspondence predictions. Meanwhile, recent advances in the functional map framework allow to enforce orientation preservation using a functional representation for tangent vector field transfer, through so-called \emph{complex functional maps}. Using this representation, we propose a new deep learning approach to learn orientation-aware features in a \emph{fully unsupervised} setting. Our architecture is built on top of DiffusionNet, making it robust to discretization changes. Additionally, we introduce a vector field-based loss, which promotes orientation preservation without using (often unstable) extrinsic descriptors. Our  code is available at: \url{https://github.com/nicolasdonati/DUO-FM}.
\end{abstract}

\section{Introduction}
\label{sec:introduction}
Learning for non-rigid shape correspondence is a key problem in 3D shape analysis with applications ranging from statistical shape analysis \cite{bogo2014faust,pishchulin2017building} to deformation or texture transfer \cite{baran2009semantic}. Early approaches have focused either on learning informative features so that corresponding points have similar feature descriptors, e.g., \cite{litman2013learning},  or modeling shape correspondence as a semantic segmentation problem. Approaches in the latter category, e.g., \cite{masci2015geodesic,monti2017geometric,poulenard2018multi,wiersma2020cnns} aim to predict, for every point on the surface, the corresponding vertex id on some ground truth template shape. Unfortunately both approaches impose very little consistency between individual point correspondence predictions, and can be sensitive to the underlying shape discretization \cite{sharp2020diffusionnet}.

\begin{figure}
\begin{center}
    \begin{overpic}
        [trim=0.0cm 0.0cm 0.0cm 0.0cm,clip,width=0.9\linewidth]{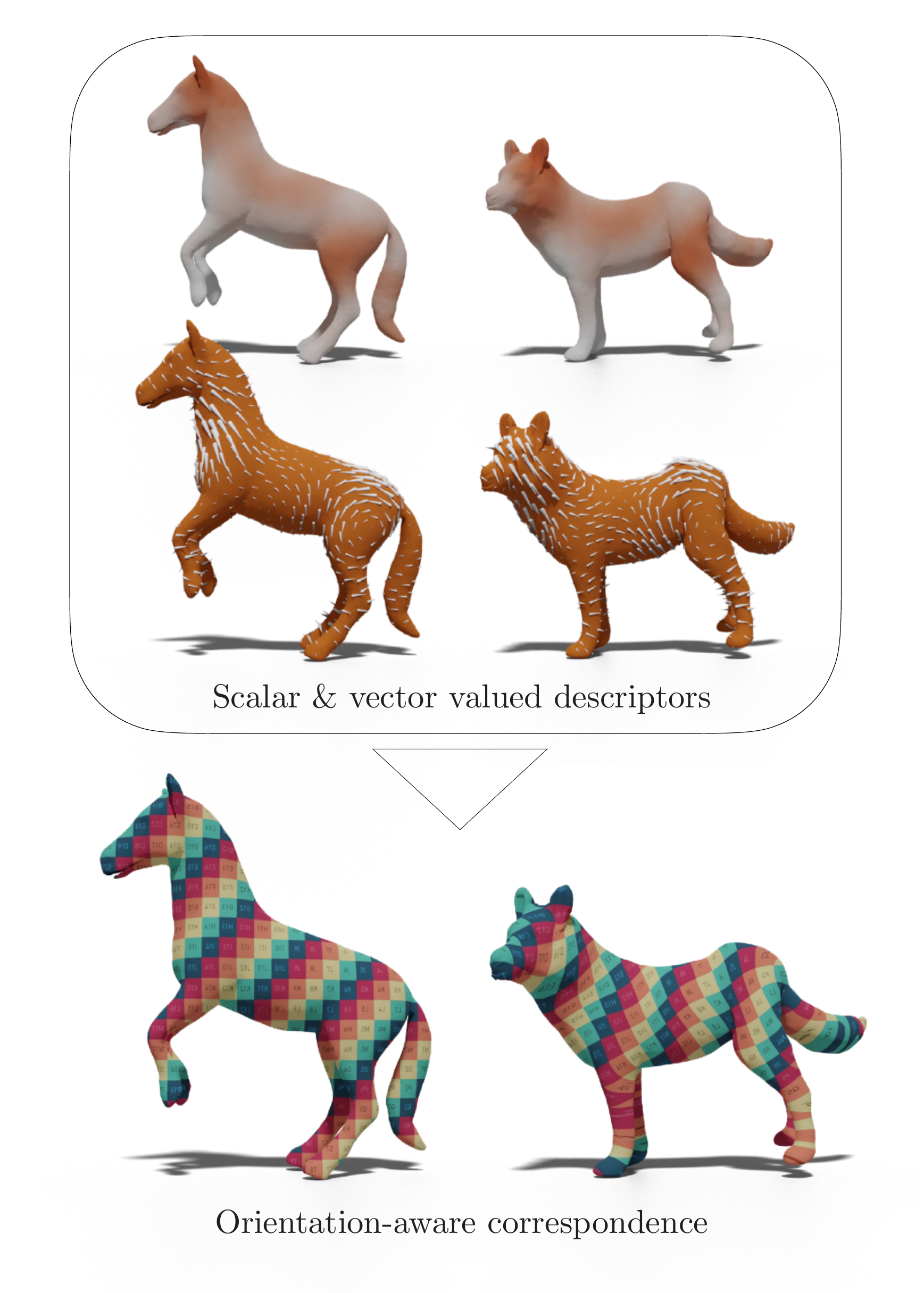}
    \end{overpic}
\end{center}
\vspace{-5mm}
\caption{Our method aims at producing \emph{orientation preserving} maps for non-rigid 3D shape matching in a fully unsupervised setting through the estimation of descriptors whose gradients also align on source and target shape.
\label{fig:teaser}}
\vspace{-3mm}
\end{figure}

More recently, techniques have focused on both predicting and imposing a training loss on the entire \emph{map} between each pair of shapes. This has been greatly facilitated by spectral approaches and especially the functional map representation \cite{ovsjanikov2012functional}, which encodes a map as a small matrix using the spectral (Laplacian) eigen-basis. A wide range of approaches based on both supervised \cite{corman2014supervised,litany2017deep,donati2020deep} and unsupervised losses \cite{roufosse2019unsupervised,halimi2019unsupervised} have been proposed using the functional map representation. Key to all of these methods is learning feature functions that are then used to predict the functional map \emph{as a whole}. As was shown across multiple recent works, this reduces the amount of necessary training data \cite{donati2020deep}, provides strong regularization promoting smooth maps,  makes the learned features robust to changes in discretization \cite{sharp2020diffusionnet}, and alleviates the requirement of the existence of a fixed template shape.

While using the compact functional map representation introduces a strong bias towards smooth approximately isometric correspondences, it nevertheless leaves room for both orientation-preserving and orientation-reversing correspondences. This orientation-agnostic property of functional maps can be useful, e.g., in symmetry detection tasks. However, in most practical scenarios, the underlying sought correspondence is expected to preserve orientation. Unfortunately, restricting to only orientation-preserving maps is not straightforward while using the functional map representation, and the maps obtained using this framework can easily introduce local and global symmetry flipping (i.e., left/right ambiguity present in many organic shapes). As a result, existing state-of-the-art learning networks require either a supervised loss \cite{litany2017deep,donati2020deep,sharp2020diffusionnet}, rigid pre-alignment \cite{sharma2020weakly}, or rely on hand-crafted extrinsic descriptors to disambiguate symmetries.

In this paper, we demonstrate that these limitations can be overcome by using the recently-proposed \emph{complex functional map} representation \cite{CompFmaps} that is based on alignment of tangent vector fields (represented as complex functions) rather than real-valued functions.

To achieve this, we propose the first architecture that uses complex functional maps and learns specific features that align tangent bundles on surfaces. The use of the complex structure makes our approach fully orientation-aware, and helps to restrict the space of allowed correspondences to only globally orientation-preserving maps, while regularizing the learning process. We introduce losses adapted to complex functional maps and demonstrate that our network can be trained in a fully unsupervised manner without relying on rigid pre-alignments or ground truth correspondences. More broadly, the vector-valued features learned by our approach provide a novel and informative signal for non-rigid shape analysis tasks.

\section{Related Works}
\label{sec:related_works}
Non-rigid shape matching is a very rich and well-established research area. Below we review works that are most closely
related to ours, focusing on learning-based, and especially unsupervised techniques. We also refer the interested
readers to surveys including \cite{biasotti2016recent,sahilliouglu2019recent} for a more in-depth overview.

\mypara{Functional Maps}
Our method builds upon the functional map representation, which was originally introduced in \cite{ovsjanikov2012functional} and then extended in a very wide range of follow-up works, e.g., \cite{eynard2016coupled,nogneng17,burghard2017embedding,wang2018vector,ren2018continuous,wang2018kernel,gehre2018interactive,Shoham2019hierarchicalFmap} among others. The key advantage of this framework is that it allows to represent and optimize for maps as small-sized matrices, enables strong linear-algebraic regularization, and can even be adapted to the partial setting \cite{rodola2017partial,litany2017fully}. 

An essential step in works using this representation, are the ``descriptor'' (also known as ``probe'' \cite{ovsjanikov2017course}) functions that are used to estimate the underlying functional maps and that must be provided \textit{a priori}. Early methods have exploited axiomatic descriptors such as heat or wave kernel signatures \cite{sun2009concise,aubry2011wave}, with several attempts aiming to optimize the weights of such descriptors through optimization techniques \cite{corman2014supervised}.

\mypara{Learning-based Methods}
Learning shape correspondence has also been done by treating it as a dense semantic segmentation problem, e.g., \cite{masci2015geodesic,boscaini2016learning,monti2017geometric,fey2018splinecnn,poulenard2018multi,wiersma2020cnns}, among many others, or \textit{via} template alignment \cite{groueix20183d}. However, such works tend to require significant amount of training data, establish a map to a template, and can fail to generalize under connectivity changes \cite{sharp2020diffusionnet}.

More closely related to our approach are methods that use learning together with the functional map representation, thus evaluating the map as a whole and allowing to directly train and test on arbitrary shape \emph{pairs}. This was first introduced in FMNet \cite{FMNet} , which proposed a method to \emph{refine} given descriptor functions such as SHOT \cite{shot} with a deep neural network, by minimizing a supervised loss, given some ground truth correspondences. This was later extended in \cite{donati2020deep}, where descriptor functions for functional map estimation are extracted directly from the shapes' geometry using point-based feature extractors, and a new regularized functional map estimation layer.

\mypara{Unsupervised Spectral Learning}
Even closer to ours are spectral approaches that use \emph{unsupervised learning} while exploiting the functional map representation. This was first done by  replacing the supervised loss in FMNet with either geodesic distance preservation \cite{halimi2019unsupervised} or desirable structural properties in the spectral domain \cite{roufosse2019unsupervised}. Other properties such as cycle consistency \cite{ginzburg2020cyclic} or unsupervised alignment of heat kernels \cite{aygun2020unsupervised} have also been used to improve efficiency and accuracy.

These methods are attractive since they do not rely on manual supervision, are fully intrinsic and thus tend to generalize well across pose changes. At the same time, their fully intrinsic nature can cause ambiguities in the presence of intrinsic symmetries such as those present in human shapes. To alleviate this problem, previous functional maps methods, typically only \emph{refine} given descriptors such as SHOT, which carry some extrinsic information \cite{halimi2019unsupervised,roufosse2019unsupervised, ginzburg2020cyclic,aygun2020unsupervised} but can unfortunately be highly unstable under connectivity changes.
More recently, ``weak supervision'' was advocated in the form of rigid pre-alignment \cite{sharma2020weakly, eisenberger2021neuromorph} to resolve symmetry ambiguity. Finally, Deep Shells \cite{eisenberger2020deep} performs SHOT feature refinement \emph{jointly} with using the 3D embedding to guide unsupervised correspondence learning.

Unfortunately, despite significant effort, symmetry ambiguity remains a central problem in unsupervised learning for non-rigid shape matching. This is especially problematic since spectral methods tend to generalize much better to unseen poses compared to extrinsic methods such as \cite{groueix20183d}. As we argue in this paper, the symmetry ambiguity problem is inherent to the functional maps approaches, as the losses used are fully intrinsic and thus cannot disambiguate orientation-preserving vs. orientation-reversing maps.

\mypara{Complex Functional Maps} Recently a tool was introduced for geometry processing, called \emph{complex functional maps} \cite{CompFmaps} aimed at alignment of \emph{tangent vector fields} rather than functions. Crucially, complex functional maps allow to remove orientation reversing maps from the space of allowed correspondences. As a result, as shown in \cite{CompFmaps}, this can help to gain better control on both orientation, and ultimately symmetry in the computed maps. However, the approach in  \cite{CompFmaps} still uses either axiomatic descriptors or an iterative approach in its pipeline. Therefore, it is unclear how to incorporate this representation into a learning framework, while maintaining accuracy and efficiency.

\paragraph*{Contributions}
Our main contributions are as follows:
\begin{enumerate}
\setlength\itemsep{0em}
\item We introduce a new \emph{orientation-aware unsupervised loss}, using the recently proposed complex functional maps representation \cite{CompFmaps}, that exploits the properties of tangent vector fields. 
\item We show that computing complex functional maps directly from gradients of learned features and then imposing an additional loss on these maps helps to regularize currently unstable pipelines with respect to symmetry aliasing.
\item By building upon a recent, robust feature extraction backbone \cite{sharp2020diffusionnet}, we introduce a fully unsupervised correspondence learning approach, without using extrinsic descriptors or coordinate information, while being robust to significant changes in triangulation. 
\end{enumerate}

\section{Background and Motivation}
\label{sec:background}
\subsection{Notation, Background \& Motivation}

Given a pair of non-rigid shapes, $M, N$, represented as triangle meshes, our main goal is to estimate a map $\varphi: M \rightarrow N$ in an unsupervised manner.

\paragraph*{Functional Maps}
In this work, we use the functional map framework, which has recently given rise to state-of-the-art supervised \cite{litany2017deep,donati2020deep,sharp2020diffusionnet} and unsupervised \cite{roufosse2019unsupervised,halimi2019unsupervised,eisenberger2020deep,ginzburg2020cyclic} learning-based non-rigid shape correspondence methods.

The key idea behind the functional maps approaches is that any correspondence can be represented compactly as a small-sized matrix. Specifically, given any map $\varphi: M \rightarrow N$, it can first be encoded as a binary matrix $\Pi_{MN}$ s.t. $\Pi_{MN}(i,j) = 1$ if and only if $\varphi(i) = j$ where $i$ and $j$ are vertices on $M$ and $N$ respectively. The associated \emph{functional map} $\mathbf{C}_{NM}$ is given as $\mathbf{C}_{MN} = \Phi_M^{\dagger} \Pi_{MN} \Phi_N$ where $\Phi_M, \Phi_N$ are matrices storing as columns the first $k$ eigenfunctions of the Laplace-Beltrami operators of shapes $M, N$, while $\dagger$ is the Moore-Penrose pseudo-inverse. Note that $\mathbf{C}_{MN}$ is of size $k \times k$ with $k$ typically between $20-100$, and is thus orders of magnitude smaller than $\Pi_{MN}$.

In addition to allowing to \emph{represent} any map in a reduced basis, the functional map representation also allows to \emph{recover} the underlying map $\Pi_{MN}$ by exploiting $\mathbf{C}_{MN}$. Here and throughout the paper we adopt the notation from \cite{ovsjanikov2017course} where objects in the reduced basis are denoted in bold.

The basic pipeline for map recovery, introduced in \cite{ovsjanikov2012functional}, assumes the presence of some descriptor functions that are expected to be preserved under the unknown mapping. If $\mathbf{A}_{N}, \mathbf{A}_{M}$ are the coefficients of descriptors in the basis $\Phi_M$ and $\Phi_{N}$, the optimal functional map $\mathbf{C}_{MN}$ is computed as:
    \begin{align}
        \label{eq:fmaps}
        \min_{\mathbf{C}_{NM}} \| \mathbf{C}_{NM} \mathbf{A}_{N}  -  \mathbf{A}_{M} \| + \alpha E_{reg}(\mathbf{C}_{NM}).
    \end{align}
Here the first term promotes preservation of descriptor functions, whereas the second is a regularizer that promotes structural properties; e.g., $E_{reg}(\mathbf{C}_{NM}) = \|\mathbf{C}_{NM} \mathbf{\Delta}_{N} - \mathbf{\Delta}_{M} \mathbf{C}_{NM} \|$, where $\mathbf{\Delta}_{M}, \mathbf{\Delta}_{N}$ are diagonal $k \times k$ matrices of Laplacian eigenvalues.

The final point-to-point map $\varphi: M \rightarrow N$ can be extracted via nearest neighbor search between the rows of $\Phi_{M} \mathbf{C}_{NM}$ and those of $\Phi_{N}$ \cite{pai2021fast}. We refer to \cite{ovsjanikov2017course} for an overview of the functional map representation and its extensions.

\paragraph{Unsupervised Learning with Functional Maps}
The compactness of the functional map representation $\mathbf{C}_{MN}$ implies that the optimization problem in Eq.~\eqref{eq:fmaps} reduces to a small scale least squares problem. On the other hand, the quality of the correspondence is intimately tied to the choice of the input descriptor functions. Early approaches have relied on hand-crafted features such as the Wave Kernel Signature \cite{aubry2011wave}. However, more recent methods have focused on \emph{learning} optimal features from data, first in the supervised setting \cite{corman2014supervised,litany2017deep} and recently using unsupervised or weakly supervised deep learning, \cite{roufosse2019unsupervised,halimi2019unsupervised,eisenberger2020deep,ginzburg2020cyclic,sharma2020weakly}.

Our approach is directly inspired by methods in the latter category. The general approach, shared by \emph{all} existing unsupervised or weakly supervised methods, is to train a neural network $\mathcal{F}_{\Theta}$ that, given a shape $M$ can produce a set of $d$ real-valued functions on $M$, $\mathcal{F}_{\Theta}(M) = \{f_1^M, f_2^M, ..., f^M_i\}, \text{ where } f_i^M: M \rightarrow \mathbb{R}$.

At training time, the network $\mathcal{F}_{\Theta}$ is presented with a set of \emph{pairs} of shapes $M, N$, and the extracted features $\mathcal{F}_{\Theta}(M), \mathcal{F}_{\Theta}(N)$ are used to estimate the functional map $\mathbf{C}_{NM}$ by first projecting the features onto the reduced basis and then solving the optimization in Eq.~\eqref{eq:fmaps} (typically ignoring the regularization term $E_{reg}$). The network parameters $\Theta$ are then optimized by minimizing a \emph{training loss} which penalizes some structural properties of the estimated functional map $\mathbf{C}_{NM}$.

The difference between existing methods \cite{roufosse2019unsupervised,halimi2019unsupervised,eisenberger2020deep,ginzburg2020cyclic,sharma2020weakly} lies primarily in: \textit{a)} The choice of feature extractor $\mathcal{F}_{\Theta}$ and \textit{b)} The training losses used for learning.

Our first observation is that the vast majority of existing unsupervised learning methods have a fundamental limitation in the presence of shapes with intrinsic self-symmetries. We summarize our observation in the following theorem:

\begin{theorem}
  Given a set of shapes $\{S_i\}$ that all contain an \emph{orientation reversing} isometric self-symmetry $\{T_i: S_i \rightarrow S_i\},$ s.t. $d_{S_i}(x_j, x_k) =  d_{S_i}(T_i(x_j), T_i(x_k))$, then a generic neural network $\mathcal{F}_{\Theta}$ that is trained by any of the losses introduced in \cite{roufosse2019unsupervised,halimi2019unsupervised,ginzburg2020cyclic,sharma2020weakly,aygun2020unsupervised} has at least two possible solutions that both lead to the global optimum of the loss.
 \label{thm:main}
\end{theorem}
\begin{proof}
  See the supplementary materials.
\end{proof}

In this theorem we call a neural network $\mathcal{F}_{\Theta}$ generic if it is capable of producing an \emph{arbitrary} function on the shape. An orientation-reversing self symmetry  is a map that is an intrinsic \emph{reflection} such as the left-right symmetry of human shapes, and $d_{S_i}$ is the geodesic distance on $S_i$.

A direct consequence of this theorem is that regardless of the neural network used, there must be at least two possible global optima, when training the networks using the unsupervised losses in the majority of existing works, in common settings involving symmetric shapes.

Existing methods have primarily tried to overcome this inherent limitation by restricting the power of the neural network, and training it, not from the shape geometry, but from some initial axiomatic features like the SHOT descriptors \cite{shot}. Unfortunately, as it has been observed in the past, e.g., \cite{poulenard2018multi,sharp2020diffusionnet}, and as we confirm in our extensive experiments, these descriptors are highly sensitive to the triangle mesh structure. Alternatively, some approaches \cite{sharma2020weakly} have relied on pre-aligning the shapes in 3D space (and enforcing a consistent forward direction) or used correspondences in 3D space to guide learning \cite{eisenberger2020deep}. Such an approach, while useful for some categories, can be difficult to enforce for arbitrary non-rigid 3D shapes.

Finally, several solutions to intrinsic orientation problem have been proposed within the functional map framework. However, most of them are descriptor-based \cite{ren2018continuous,shot} which are unreliable, and form very weak constraints: the map does not have to be orientation preserving but instead encouraged to follow (possibly noisy) descriptors.

\subsection{Complex Functional Maps}
\label{sec:backg_QMaps}

In this work, we propose to address the limitations mentioned above by exploiting the recently-proposed \emph{complex functional map} representation \cite{CompFmaps}. The fundamental observation leading to this new tool is that it is challenging to recover orientation, a global signal, from a point-to-point mapping giving only local information. Therefore, the construction in \cite{CompFmaps} relies on global analysis of the \emph{pushforward} $\diff \varphi : TM \rightarrow TN$ associated to $\varphi$ whose local properties are related to normal orientation. By definition, a pushforward maps tangent vectors at $p \in M$ to tangent vectors at $\varphi(p) \in N$.
$\diff \varphi$ is also called the differential of $\varphi$ and is the best linear approximation of the map at point $p$.

A complex functional map $Q : TM \rightarrow TN$ is a relaxed representation of the pushforward in the sense that it maps any tangent vector field on $M$ to a tangent vector field on $N$ with only one constraint: it must be \emph{complex linear}: $Q(zX) = zQ(X), z\in \C, X \in TM$. The adjective complex comes from the fact that tangent vector fields are represented by complex valued functions as in \cite{Sharp:2019:VHM}.

By definition, the pushforward contains the information of pointwise mapping (which tangent planes on $N$ correspond to tangent planes on $M$) already carried by $\varphi$ thus $C$ and $Q$ cannot be independent. Moreover, $Q$ also contains local orientation information related to the orientation of the outward normals. As shown in \cite{CompFmaps}, since $Q$ is complex linear it can only represent \emph{orientation preserving} maps. These fundamental properties have been summarized in Thm. \ref{thm:Q_pushforward} proved in \cite{CompFmaps} (Sec. 3.5).
\begin{theorem}
    The complex-linear map $Q$ is a pushforward if and only if there exists an \emph{orientation-preserving} and conformal diffeomorphism $\varphi : M \rightarrow N$ satisfying:
	\begin{equation}
	    \langle X, \nabla (f \circ \varphi) \rangle_{T_pM} = \langle Q (X), \nabla f \rangle_{T_{\varphi(p)}N} ,
	\label{eq:defdT}
	\end{equation}
    for all $X \in TM, f \in L^2(N), p \in M$.
\label{thm:Q_pushforward}
\end{theorem}

Apart from this complex-linearity, the construction of complex functional maps (which we also call $Q$-maps) is analogous to standard functional maps described above. They can be written in the spectral basis $\lbrace \Psi^i \rbrace_{i\in(1,k)}$ of the connection Laplacian $\mathbf{L}$ introduced in~\cite{Sharp:2019:VHM}. In these reduced spaces, $Q$-maps are small matrices transferring coefficients in the basis $\mathbf{\Psi}_M$ to coefficients in $\mathbf{\Psi}_N$. A point-to-point map is extracted from a nearly isometric $\mathbf{Q}$ using a nearest-neighbor search on Dirac functions: $\mathbf{\Pi}_{MN} = \mathrm{NNsearch}(\mathbf{\divg}_M \mathbf{\Psi}_M, \mathbf{\divg}_N \mathbf{\Psi}_N \mathbf{Q})$ (see \cite{CompFmaps}, Section 4.6).

Moreover, a pushforwad $\mathbf{Q}$ is isometric if and only if it commutes with the connection Laplacian: $\mathbf{Q} \mathbf{L}_M = \mathbf{L}_N \mathbf{Q}$.  Finally, as shown in Sec. 3.6 of~\cite{CompFmaps} a necessary condition for $\mathbf{Q}$ to represent a pushforward is that $\mathbf{Q}$ must be orthogonal \textit{i.e.} $\mathbf{Q}^\star \mathbf{Q} = \mathbf{I}$ where ${ }^\star$ denotes the complex transposition. Here, unlike functional maps, orthogonality is not equivalent to area-preservation.

A complex functional map can be estimated by minimizing a simple optimization problem, similar to Eq.~\eqref{eq:fmaps}:
\begin{equation}
    \mathbf{Q}_{MN} = \argmin_\mathbf{Q} \| \mathbf{Q} \mathbf{B}_M - \mathbf{B}_N \|_F^2 + E_{\text{reg}}(\mathbf{Q}),
\label{eq:Q_estimation}
\end{equation}
where $E_{\text{reg}}(Q) = w_{\text{Q-ortho}} \| \mathbf{Q}^\star \mathbf{Q} - \mathbf{I} \|_F^2 + w_{\text{Q-iso}} \| \mathbf{Q} \mathbf{L}_M - \mathbf{L}_N \mathbf{Q} \|_F^2$, and $\mathbf{B}$ are the coefficients of complex (tangent vector)-valued features expressed in spectral basis of the corresponding shape.


\section{Method}
\label{sec:method}
\begin{figure}
\begin{center}
    \begin{overpic}
        [trim=0.0cm 0.0cm 0.0cm 0.0cm,clip,width=1\linewidth]{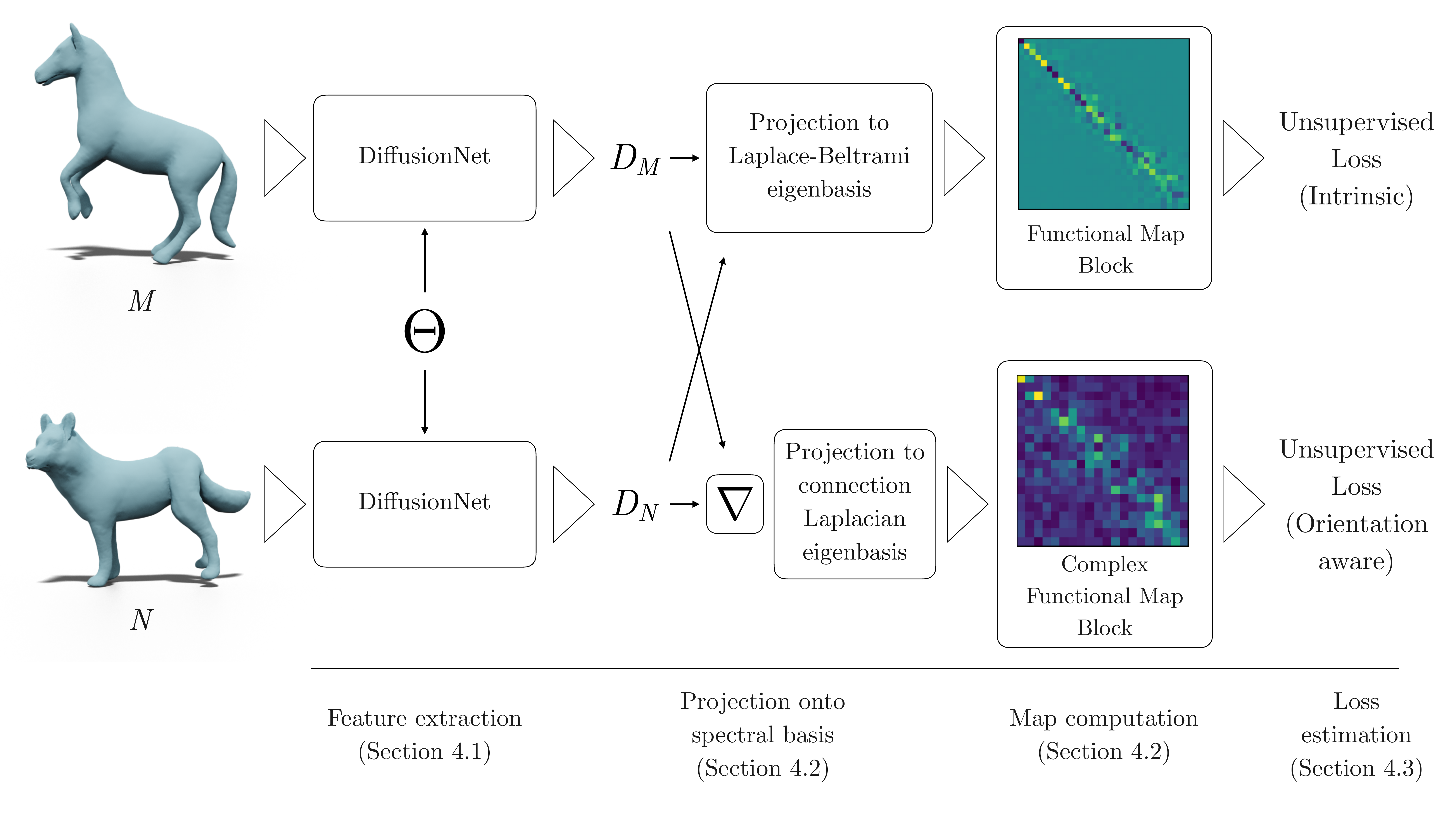}
    \end{overpic}
\end{center}
\vspace{-3mm}
\caption{Overview of our unsupervised network. We extract source and target descriptors $D_M$ and $D_N$ using DiffusionNet \cite{sharp2020diffusionnet} and then project descriptors onto the Laplace-Beltrami eigenbasis and descriptor gradients onto the connection Laplacian eigenbasis. This leads respectively to the Functional Map block (Section \ref{sec:CFMap}) and the Complex Functional Map block (Section \ref{sec:QFMap}). Losses are imposed on both of these maps (Section \ref{sec:losses}). 
\label{fig:method}}
\end{figure}

In this section, we describe our proposed network in detail. As mentioned in Section \ref{sec:background}, a deep functional map pipeline can be decomposed into three different building blocks: the feature extractor (Sec.~\ref{sec:feature_extractor}), the non-learnable functional map layer (Sec.~\ref{sec:Fmap_blocks}) and the loss (Sec.~\ref{sec:losses}). We describe our design choices for each of these components, and how they permit orientation-aware unsupervised learning in the subsections below. We also provide a graphic representation of our whole approach in Figure~\ref{fig:method}.

\subsection{Feature Extractor}
\label{sec:feature_extractor}

The first major component of our network is the deep feature extractor. Its structure is that of a Siamese network extracting features for both source and target shapes. We use a recent surface feature extractor backbone DiffusionNet \cite{sharp2020diffusionnet}, and use the Wave Kernel Signature \cite{aubry2011wave} (WKS) as input for the network, for its rotational invariance property. DiffusionNet then outputs feature vectors of dimension $d$ on the source and target shapes (respectively composed of $n_M$ and $n_N$ vertices). We denote by $D_M, D_N \in \mathbb{R}^{n_M \times d} \times \mathbb{R}^{n_N \times d}$ the learned source and target features.

The use of DiffusionNet \cite{sharp2020diffusionnet} makes our approach highly robust to changes in \emph{shape triangulation} unlike previous unsupervised approaches relying on SHOT descriptors~\cite{shot}. These methods usually fail if trained and tested triangulations are different, as demonstrated in Section \ref{sec:results}.
On the contrary, DiffusionNet is based on robust diffusion, and is consequently largely independent on the choice of shape triangulation. 
Although learned diffusion is fully intrinsic, the network is aware of the shape orientation because of oriented gradient blocks, as described in \cite{sharp2020diffusionnet} (Section 3.4). Therefore, our feature extractor can produce orientation-aware features that we use later to estimate orientation-preserving complex functional maps.

To sum up, we use DiffusionNet jointly with WKS inputs to build discretization-agnostic features. In the results Section \ref{sec:results}, we exhibit shapes with anisotropic triangulations to illustrate that methods relying on SHOT descriptor refinement tend to overfit to the triangulation rather than learn relevant shape information.

\subsection{The Functional Map Blocks}
\label{sec:Fmap_blocks}

This block, first introduced in \cite{FMNet} estimates the functional map in a differentiable way from the source and target features estimated by the feature extractor. 

\subsubsection{Regularized Functional Map Block}
\label{sec:CFMap}
The input features $D$ are projected on spectral Laplace-Beltrami eigenbasis $\Phi_S$ to get spectral features $\mathbf{A}_S = \Phi_S^{\dagger} D_S$ with $S \in \{M, N\}$. The functional map $\mathbf{C}_{NM}$ is then estimated as the solution to the following least-squares problem
: 
\begin{align*}\mathbf{C}_{NM} = \argmin_{\mathbf{C}} \| \mathbf{C}   \mathbf{A}_N - \mathbf{A}_M \|_F^2,
\end{align*}
leading to:
\begin{equation}
    \mathbf{C}_{NM} = \mathbf{A}_M \mathbf{A}_N^{\dagger} .
\end{equation}
In this work, we use the \textit{regularized} approach, introduced in  \cite{donati2020deep}, which incorporates the Laplacian commutativity energy $\|\mathbf{C}_{NM} \mathbf{\Delta}_{N} - \mathbf{\Delta}_{M} \mathbf{C}_{NM} \|_F^2$ in a differentiable manner, directly in the functional map estimation step.

\subsubsection{Complex Functional Map Block}
\label{sec:QFMap}
The complex functional map estimation is analogous to that of the standard functional map. We first convert the feature functions $D$ to vector fields using the discrete gradient operator $G$. We visualize these vector field descriptors in Figure~\ref{fig:teaser} (where they are rotated by $\pi/2$ to better see singularities). These vector valued descriptors are then projected in the eigenbasis  $\Psi$ of the connection Laplacian. This leads to complex spectral feature vectors $\mathbf{B}_S = \Psi_S^{\dagger} G_S D_S$ with $S \in \{M, N\}$, and $G_S$ the gradient operator on shape $S$.

The complex functional map $\mathbf{Q}_{MN}$ is then estimated as the solution to the following least-squares problem:
$$\mathbf{Q}_{MN} = \argmin_{\mathbf{Q}} \| \mathbf{Q} \mathbf{B}_M - \mathbf{B}_N \|_F^2,$$
whose closed-form solution is given simply as:
\begin{equation}
    \mathbf{Q}_{MN} = \mathbf{B}_N \mathbf{B}_M^{\dagger}.
\end{equation}
In our work, we extend the in-network Laplacian regularization \cite{donati2020deep}, and apply it to complex functional map estimation by modifying the least squares system in the same way as was done in \cite{donati2020deep} for real-valued functional maps.

Remark that as mentioned in Section \ref{sec:backg_QMaps}, the complex functional map estimated from feature gradients is a pushforward if and only if the features themselves give rise to an orientation-preserving map. We elaborate on the relation between the two blocks in Section \ref{subsec:correlation_blocks} below. Specifically, we demonstrate that although $\mathbf{C}$ and $\mathbf{Q}$ are estimated independently, they still satisfy the equation of Thm.~\ref{thm:Q_pushforward}.

\subsection{Losses}
\label{sec:losses}
From estimated $\mathbf{C},\mathbf{Q}$ we build a loss inspired by SURFMNet \cite{roufosse2019unsupervised}. 

\paragraph*{Loss on $\mathbf{C}$.}
SURFMNet \cite{roufosse2019unsupervised} imposes the estimated functional map $\mathbf{C}$ to be orthogonal, resulting in the first loss $L_{\text{ortho}}$:
\begin{equation}
    L_{\text{ortho}}(\mathbf{C}) = \| \mathbf{C}^\top \mathbf{C} - \mathbf{I} \|_F^2
\end{equation}
Moreover, they also propose to promote isometry through commutativity between $\mathbf{C}$ and the Laplace-Beltrami operators $\mathbf{\Delta}_M, \mathbf{\Delta}_N$, resulting in the second loss $L_{\text{iso}}$:
\begin{equation}
    L_{\text{iso}}(\mathbf{C}) = \| \mathbf{C} \mathbf{\Delta}_N - \mathbf{\Delta}_M \mathbf{C} \|_F^2
\end{equation}
As stated previously, we remark that this isometric loss \emph{is not necessary} if we estimate $\mathbf{C}$ with the Laplacian regularizer of \cite{donati2020deep}. Indeed, the regularizer only produces maps that have a low isometric loss. We therefore only use $L_{\text{ortho}}$ in our implementation.

The association of these two losses is generally enough to estimate intrinsically an isometric map. The fundamental problem that we propose to remedy here is the fact that these two losses are not in themselves sufficient to rule out intrinsic symmetries. We stress again that many previous works rule out these symmetries based on triangulation only, using the SHOT feature extractor. However these methods are then biased towards the training triangulations.

\paragraph*{Loss on $\mathbf{Q}$.}
As demonstrated in \cite{CompFmaps} (Section 3.6), a complex functional map will only encode a pointwise map (which will then be orientation preserving) if it is an orthogonal matrix. Hence the complex orthogonal loss $L_{\text{Q-ortho}}$:
\begin{equation}
    L_{\text{Q-ortho}}(\mathbf{Q}) = \| \mathbf{Q}^\star \mathbf{Q} - \mathbf{I} \|_F^2
\end{equation}

Moreover, since we aim for maps which are as isometric as possible, we can also use a complex isometric loss $L_{\text{Q-iso}}$, evaluating the lack of commutativity with the connection Laplacians:
\begin{equation}
    L_{\text{Q-iso}}(\mathbf{Q}) = \| \mathbf{Q} \mathbf{L}_M - \mathbf{L}_N \mathbf{Q} \|_F^2
\end{equation}

We observe that this isometric loss can also be avoided by computing $Q$ using a Laplacian regularizer \cite{donati2020deep}. We therefore only use $L_{\text{Q-ortho}}$ in our implementation.

\paragraph*{Loss Function.}
In summary, we use $2$ losses that we combine to compute the final loss $L_{final}$:
\begin{align*}
L_{\text{final}}(\mathbf{C},\mathbf{Q}) =  w_{\text{ortho}} L_{\text{ortho}}(\mathbf{C}) + w_{\text{Q-ortho}} L_{\text{Q-ortho}}(\mathbf{Q})
\end{align*}

These losses, jointly with the Laplacian regularizers ensure that the learned descriptors result in an isometric map and that this map is orientation-preserving. The whole pipeline remains both light and unsupervised.

\subsubsection{Correlation of the Two Functional Map Blocks}
\label{subsec:correlation_blocks}
Note that we never explicitly use Eq.\eqref{eq:defdT} relating $\mathbf{C}$ and $\mathbf{Q}$ required by Theorem~\ref{thm:Q_pushforward}. However, we prove that when both functional maps are nearly isometric and estimated from the same features this relation is always verified.

\begin{theorem}
    Let $M,N$ be two manifolds, and $F_M, F_N$ surface features such that the functional map $C$ estimated from these features is an isometry. Let $Q$ be the complex functional map computed with the feature gradients as described in Section \ref{sec:QFMap}.
    Then the maps $(C,Q)$ must satisfy Eq.~\eqref{eq:defdT}, and $C$ is an orientation-preserving isometry.
    \label{thm:C_and_Q}
\end{theorem}
\begin{proof}
  See the supplementary materials.
\end{proof}

The isometric assumption is not restrictive in the sense that deep spectral methods already implicitly make this assumption. Moreover, as we demonstrate below, our approach is robust even for non-isometric shape categories.

\subsection{Implementation}
\label{sec:implementation}

We implemented our method with Pytorch 1.8 (this version is required to include complex Tensors in the differentiable pipeline) by adapting the open-source implementation of DiffusionNet \cite{sharp2020diffusionnet} for the feature extractor and \cite{donati2020deep} for the functional map block with Laplacian regularizer.

We use WKS descriptors \cite{aubry2011wave} as input signal for the network. We use this descriptor because it is: \textit{a)} Robust to changes in the shape triangulation, and captures the intrinsic geometry of the surface. As shown in the next section that guarantees that the network learns relevant surface information, rather than overfit to the mesh triangulation. \textit{b)} Independent of the embedding of the shape. This makes our approach fully rotationally invariant, and capable of predicting correspondences between arbitrarily rotated shapes. Indeed methods such as \cite{sharma2020weakly, eisenberger2021neuromorph} depend on pre-aligned datasets to work, which makes them only weakly supervised instead of fully unsupervised.

Our feature extraction network consists of 4 DiffusionNet blocks (a standard DiffusionNet setup \cite{sharp2020diffusionnet}), where the 128-dimensional input WKS features are transformed by each block to learned features of same dimension 128, to finally produce 128-dimensional descriptors on source and target shape.
As described in Section \ref{sec:feature_extractor}, our network is applied in a Siamese way on the two input shapes, using the same weights for feature extraction on source and target.

\vspace{-2mm}\paragraph*{Parameters}
\label{parameters}

In addition to the architecture above, our method has some key hyper-parameters: \textit{a)} The size of both spectral basis: we use $k_C=50$ for Laplace-Beltrami and $k_Q=20$ for connection Laplacian \textit{b)} The Laplacian regularizer from \cite{donati2020deep} in the functional map blocks: we use $\lambda = 10^{-3}$ as recommended in the original paper \textit{c)} The loss hyper-parameters:
The loss if focused on map orthogonality since Laplacian-commutativity is enforced previously with the regularizers.
We enforce both maps to be ``equally" orthogonal,
by setting $w_{\text{ortho}} = w_{\text{Q-ortho}} = 1$.

We train our network with a batch size of 1 for a number of epochs between 5 and 30. We use a learning rate of $10^{-3}$ with ADAM optimizer \cite{adam}.

\section{Results}
\label{sec:results}
\begin{table}
\begin{center}
\newrobustcmd{\B}{\bfseries}

\begin{tabular}{|l| p{.85cm} p{.85cm} | p{.85cm} p{.85cm} |}

\hline

\B Meth / Data &F\_r/F\_r&F\_r/F\_a&S\_r/S\_r&S\_r/S\_a\\
\hline\hline
SHOT+FMN \cite{FMNet}
&  5.8    &  43.    &  7.0    &  41.    \\
WKS+GFM \cite{donati2020deep}
&\B2.0    &\B2.6    &\B2.2    &\B2.3    \\
\hline
\hline
BCICP \cite{ren2018continuous}
&  6.1    &  8.5    &  11.    &  14.    \\
ZO \cite{zoomout}
&  6.1    &  8.7    &  7.5    &  15.    \\
\hline
SHOT+UFMN \cite{halimi2019unsupervised}
&  5.7    &  42.    &  9.9    &  44.    \\
SHOT+DS \cite{eisenberger2020deep}
&\B1.7    &  12.    &\B2.5    &  10.    \\
WKS+DS \cite{eisenberger2020deep}
&  8.2    &  9.5    &  8.3    &  20.    \\
\B WKS+Ours    &  2.5    &\B3.0    &  2.6    &\B2.7    \\
\hline
\hline
\multicolumn{5}{|c|}{Cross Training}            \\
\hline
\B Meth / Data &S\_r/F\_r&S\_r/F\_a&F\_r/S\_r&F\_r/S\_a\\
\hline\hline
SHOT+FMN \cite{FMNet}
&  14.    &  43.    &  11.    &  44.    \\
WKS+GFM \cite{eisenberger2020deep}
&\B9.9    &\B8.4    &\B3.8    &\B3.9    \\
\hline
\hline
SHOT+UFMN \cite{halimi2019unsupervised}
&  12.    &  44.    &  9.3    &  43.    \\
SHOT+DS \cite{eisenberger2020deep}
&\B2.7    &  15.    &  5.7    &  16.    \\
WKS+DS \cite{eisenberger2020deep}
&  6.7    &  12.    &  9.2    &  21.    \\
\B WKS+Ours    &\B2.7    &\B3.1    &\B4.2    &\B4.4    \\

\hline

\end{tabular}
\end{center}
\vspace{-3mm}
\caption{Comparative results ($\times 100$) of all main baselines on FAUST and SCAPE re-meshed and anisotropic. Deep Learning methods are displayed with the descriptor input they were fed during training time. The methods shown in the top group are supervised, while the ones below (separated by a double line) are axiomatic or unsupervised. Note that our approach outperforms all unsupervised baselines \emph{without post-processing}, and achieves similar or better performance even to supervised ones.\vspace{-1mm}}
\vspace{-3mm}
\label{table:aniso}
\end{table}

In this section, we show that our network can outperform state-of-the-art deep shape matching architectures on standard datasets like FAUST (F\_r) \cite{bogo2014faust} and SCAPE (S\_r) \cite{anguelov2005scape} as-well-as non-isometric datasets like SHREC'19 \cite{SHREC19} and SMAL \cite{Zuffi2017smal}. Following \cite{ren2018continuous}, all shapes are remeshed so that they do not share the same connectivity. Moreover, we introduce an anisotropic re-meshing of FAUST (denoted F\_a) and SCAPE (denoted S\_a), generated with Mmg \cite{dobrzynski2008anisotropic,dapogny2014three}, to demonstrate how some methods overfit to mesh connectivity to disambiguate between intrinsic symmetries.
We show our anisotropic remeshings in the supplementary materials. 

In all our tables, we denote by F/S a method trained on the dataset F and tested on S.

\subsection{Quantitative Results}

\paragraph*{Baselines}

We compare our method to:
\begin{itemize}
    \item Axiomatic methods: BCICP \cite{ren2018continuous} and ZoomOut \cite{zoomout} are very efficient for solving close to isometric matching. It should also be noted that these axiomatic methods are slower than a test pass of our method \emph{which does not require post-processing}.
    
    \item Supervised methods: FMNet \cite{FMNet} (denoted as FMN) and GeoFmaps \cite{donati2020deep} (denoted as GFM), using DiffusionNet as feature extractor.
    
    \item Unsupervised methods: Unsupervised-FMNet \cite{halimi2019unsupervised} (denoted as UFMN), and the state-of-the-art method Deep Shells \cite{eisenberger2020deep} (denoted as DS).
\end{itemize}

For the most relevant baselines, we compare our method, which uses WKS descriptors, to both original networks (trained with SHOT descriptors) and their variant when trained with WKS as input. Consequently, we denote as ``SHOT+Net" a Network trained with SHOT and ``WKS+Net" its variant with WKS as input.

\begin{table}
\begin{center}
\newrobustcmd{\B}{\bfseries}

\begin{tabular}{|l|l l|l|}
\hline
Meth / Data    & F\_r/Sh\_r&S\_r/Sh\_r&Sh\_r/Sh\_r\\
\hline\hline
BCICP \cite{ren2018continuous}
&  15.      &  15.     &  15.      \\
ZO \cite{zoomout}
&  21.      &  21.     &  21.      \\
\hline
SHOT+DS \cite{eisenberger2020deep}
&  27.      &  24.     &  24.      \\
WKS+DS  \cite{eisenberger2020deep}
&  27.      &  29.     &  28.      \\
\B WKS+Ours    &\B6.4      &\B8.4     &\B3.9      \\
\hline
\end{tabular}
\end{center}
\vspace{-3mm}
\caption{Comparative results ($\times 100$) on SHREC'19 re-meshed with different train sets, including SHREC'19 re-meshed itself. We compare unsupervised methods on this more challenging dataset, and keep the same notations as in Table \ref{table:aniso}. We see that our approach gives the best correspondences and their quality is relatively stable with respect to the training set. 
}
\label{table:shrec}
\vspace{-3mm}
\end{table}

\paragraph*{Anisotropic FAUST and SCAPE}

For this experiment we train networks on FAUST and SCAPE re-meshed as in \cite{ren2018continuous}, and test them on both re-meshed and anisotropic. We report the mean geodesic errors in Table \ref{table:aniso}.

From the results shown in Table \ref{table:aniso}, we see that:
\textit{a)} Our method is robust to triangulation changes which make SHOT-based methods fail at test time (often because they mistake anisotropy for meaningful geometric information).   Deep Shells \cite{eisenberger2020deep} is the most robust SHOT-based approach, since its feature extractor uses spectral filters which helps filter out high-frequency overfitting. Still, the quality of the correspondence collapses on anisotropic datasets.
\textit{b)} Deep Shells, if presented with an intrinsically symmetric signal (here WKS \cite{aubry2011wave}) as input, fails to learn accurate descriptors, resulting in overall poor correspondence.
\textit{c)} Our approach gives the best results among unsupervised networks. Moreover, the quality of the correspondence is often close to that achieved by the best supervised baseline \cite{donati2020deep}. 

\paragraph*{SHREC'19 Re-meshed}

For this second experiment, we train the networks respectively on FAUST, SCAPE and SHREC'19 (denoted as F\_r, S\_r and Sh\_r) re-meshed so that mesh connectivity are different and tested only on SHREC'19.
We removed shape 40 from SHREC'19 in this experiment since it is the only partial non-closed shape and therefore outside the scope of this method.

We show in Table \ref{table:shrec} that Deep Shells fails to generalize to this test set, even though all meshes have similar number of vertices and well-shaped meshes. 
In comparison, our method gives significantly better results, as it is a purely geometric approach tailored to exploit surface information rather than triangulation details and to produce well-oriented maps.

\begin{table}
\begin{center}
\newrobustcmd{\B}{\bfseries}

\begin{tabular}{|l|l|}
\hline
Meth / Data   & SMAL\_r \\
\hline\hline
BCICP \cite{ren2018continuous}
&  19.    \\
ZO \cite{zoomout}
&  35.    \\
\hline
SHOT+DS \cite{eisenberger2020deep}
&  25.    \\
WKS+DS \cite{eisenberger2020deep}
&  33.    \\
\B WKS+Ours   &\B4.8    \\
\hline
\end{tabular}
\end{center}
\vspace{-3mm}
\caption{Comparative results ($\times 100$) on SMAL re-meshed dataset. This animal dataset exhibits strong non-isometries, as can be seen on the qualitative result in Figure \ref{fig:smal_quali}, to which only our method proves to be robust.
}
\label{table:smal}
\vspace{-3mm}
\end{table}

\paragraph*{SMAL Re-meshed}

We test unsupervised methods on SMAL shapes \cite{Zuffi2017smal} (again re-meshed so that connectivity is different on every mesh as in \cite{ren2018continuous}). The dataset, originally composed of 49 shapes, is split in 32 training shapes and 17 test shapes. This dataset constitutes the hardest of the three experiments, since its shapes are \textit{often strongly non-isometric.} Indeed, they involve animal shapes of different species, which often poses significant challenges for existing (especially spectral) approaches. 

The results of this experiment are reported in Table \ref{table:smal}. Observe that our method is the \textit{only} one that produces even reasonable correspondence, significantly outperforming the closest competitor. This highlights the significant additional robustness ensured by our combination of an accurate feature extractor with a range of well-founded geometric regularizers, which allow our approach to accommodate even for non-isometric shapes and learn from limited training data.

\subsection{Qualitative Results}

We additionally provide a qualitative result for our third experiment in Figure \ref{fig:smal_quali}, comparing our method with baselines on SMAL. We see that our method yields a map very close to ground-truth, even on this challenging example with strong non-isometric distortions. Meanwhile, both axiomatic and learning baselines fail to predict accurate correspondences.
We provide another qualitative comparison in the supplementary materials.

\begin{figure}
\begin{center}
\includegraphics[width=1\linewidth]{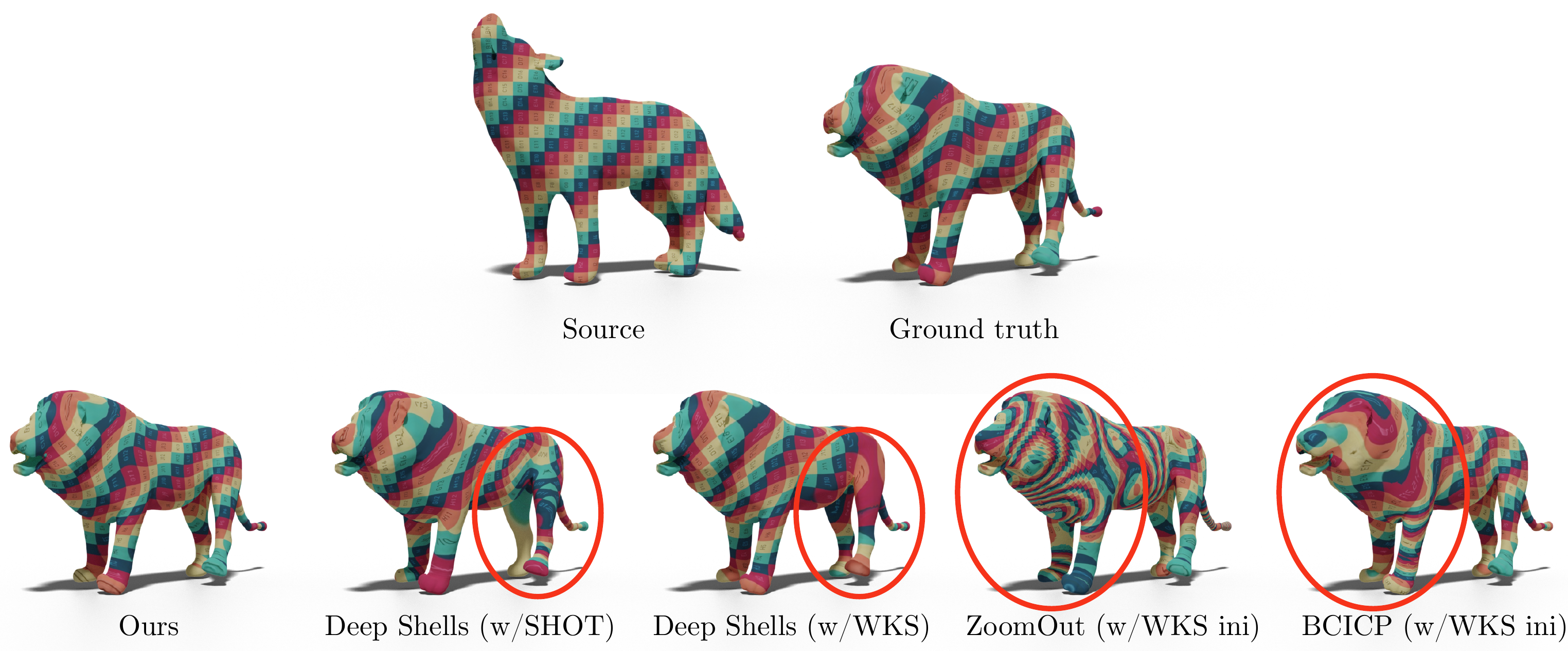}
\end{center}
\vspace{-2mm}
\caption{Qualitative results for baselines on the SMAL dataset. The areas where baselines gave the most wrong predictions are highlighted in red.
\label{fig:smal_quali}}
\vspace{-2mm}
\end{figure}

\section{Conclusion, Limitations \& Future Work}
\label{sec:conclusion}
To conclude, we introduced a new fully unsupervised way to efficiently learn accurate descriptors for shape matching. These descriptors are guaranteed by our complex functional map loss to be orientation aware in the sense that the resulting map is orientation preserving, making the pipeline robust to aliasing due to shape intrinsic symmetries. Besides, the use of DiffusionNet for a feature extractor enables robustness to changes in shape discretization.

Our approach has several limitations: the loss, which aims at as-isometric-as-possible functional maps is dependent on the fact that the input shapes are not too non-isometric. It would therefore be interesting to use spectral bases adapted to non-isometry, as in \cite{rakoto2020learnembed}, which we leave as future work.
Another current limitation of our method is that it needs manifold meshes as input. However, good Laplacian operators can be built on point clouds \cite{Sharp:2020:LNT}, and it would be interesting to check how robust our pipeline is to potentially noisy point cloud inputs.
Finally, we believe it would be interesting to further leverage complex functional maps in other learning applications, while promoting general rotation-invariant and orientation-preserving maps.

\mypara{Acknowledgements}
Parts of this work were supported by the ERC Starting Grant No. 758800 (EXPROTEA) and the ANR AI Chair AIGRETTE.

{\small
\bibliographystyle{ieee_fullname}
\bibliography{egbib.bib}
}

\newpage
\appendix

\section{Proof of Theorem \ref{thm:main}, \ref{thm:Q_pushforward}, \ref{thm:C_and_Q}}

\paragraph{Theorem 1.}
\textit{
\hspace{-4mm}
Given a set of shapes $\{S_i\}$ that all contain an \emph{orientation reversing} isometric self-symmetry $\{T_i: S_i \rightarrow S_i\},$ s.t. $d_{S_i}(x_j, x_k) =  d_{S_i}(T_i(x_j), T_i(x_k))$, then a generic neural network $\mathcal{F}_{\Theta}$ that is trained by any of the losses introduced in \cite{roufosse2019unsupervised,halimi2019unsupervised,ginzburg2020cyclic,sharma2020weakly,aygun2020unsupervised} has at least two possible solutions that both lead to the global optimum of the loss.
}

\begin{proof}
    The spectral losses $L$ defined in~\cite{roufosse2019unsupervised,halimi2019unsupervised,ginzburg2020cyclic,sharma2020weakly,aygun2020unsupervised} are fully intrinsic, thus they are invariant under shape isometric changes \textit{i.e.} $L \circ T_i = L$. So, if all shapes admit an isometric self-symmetry, the solution composed with the isometry will have the same loss value.
\end{proof}

\paragraph{Theorem 2.}
\textit{
\hspace{-4mm}
The complex-linear map $Q$ is a pushforward if and only if there exists an \emph{orientation-preserving} and conformal diffeomorphism $\varphi : M \rightarrow N$ satisfying:
\begin{equation}
    \langle X, \nabla (f \circ \varphi) \rangle_{T_pM} = \langle Q (X), \nabla f \rangle_{T_{\varphi(p)}N},
\label{eq:defdT}
\end{equation}
for all $X \in TM, f \in L^2(N), p \in M$.
}

\begin{proof}
    See Theorem 3.1 in \cite{CompFmaps}, Section 3.5
\end{proof}

\paragraph{Theorem 3.}
\textit{
\hspace{-4mm}
Let $M,N$ be two manifolds, and $F_M, F_N$ surface features such that the functional map $C$ estimated from these features is an isometry. Let $Q$ be the complex functional map computed with the feature gradients as described in the main manuscript.
Then the maps $(C,Q)$ satisfy Eq.~\eqref{eq:defdT}, and $C$ is an orientation-preserving isometry.
}

\begin{proof}
    By assumption the functional map $C : L^2(N) \rightarrow L^2(M)$ represents the isometric map $\varphi : M \rightarrow N$ and exactly transfers the descriptors \ie $C(F_N) = F_M$. Moreover the complex functional map $Q : TM \rightarrow TN$ transfers the gradient of the descriptors $Q(\nabla_M F_M) = \nabla_N F_N$ and is complex-linear.
    
    To recover Eq.~\eqref{eq:defdT}, we take the inner product of the gradient transfer with the gradient of an arbitrary function $f : N \rightarrow \R$: 
    \begin{align*}
        \I^N_p \left( Q( \nabla_M F_M), \nabla_N f \right) &= \I^N_p \left( \nabla_N F_N, \nabla_N f \right).
    \end{align*}
    
    This equation easily simplifies using the properties of an isometric map: the metric is preserved by the pullback ($\varphi^\star \I^N = \I^M$) and the pushforward commutes with the gradient ($\diff\varphi^{-1} \left(\nabla_N f \right) = \nabla_M C(f)$), yielding:
    \begin{align*}
        &\I^N_p \left( Q( \nabla_M F_M), \nabla_N f \right) \\
            &= \left(\left(\varphi^\star\right)^{-1} \I^M\right)_{\varphi^{-1}(p)} \left( \diff\varphi^{-1} \left(\nabla_N F_N\right), \diff\varphi^{-1} \left(\nabla_N f \right)\right) \\
            &= \I^M_{\varphi^{-1}(p)} \left( \nabla_M C(F_N), \nabla_M C(f)\right)
    \end{align*}
    
    So $Q$ and $C$ satisfy Eq.~\eqref{eq:defdT} for all complex-linear combination of the gradient descriptors. Therefore, following Thm.~\ref{thm:Q_pushforward}, $Q$ is the pushforward associated to $\varphi$ and $C$ must be orientation preserving.
\end{proof}

\section{Ablation Study}

\begin{table}
\begin{center}

\begin{tabular}{|l|c|}
\hline
Meth / Data           & SMAL\_r \\
\hline\hline
xyz input-3 axis      &  25.    \\
xyz input-1 axis      &  5.9    \\
\hline
nonOA-FE              &  34.    \\
\hline
no $Q$-maps (epoch 3) &  5.8    \\
no $Q$-maps (epoch 15) &  8.1    \\
\hline
\textbf{Ours} (epoch 3)  &\textbf{4.8}    \\
Ours (epoch 15)  & 5.1    \\
\hline
\end{tabular}
\end{center}
\caption{Comparative results ($\times 100$) for the different ablations of our method.}
\label{table:ablation_study}
\end{table}

This section presents an ablation study for the most vital components of our approach, namely: \textit{a)} The input signal fed to the network, \textit{b)} The orientation-aware feature extractor, \textit{c)} The orientation-aware loss.
We test these ablations on our third experiment, on the SMAL dataset \cite{Zuffi2017smal} (see main manuscript, Section 5.1 for more details), and to a maximum of 20 epochs. We compare these three ablations to our approach and report the results in Table \ref{table:ablation_study}. The ablations are commented in details in the sections below.

\subsection{The WKS Descriptors as Input Features}
    As stated in the main manuscript, many unsupervised deep learning for non-rigid 3D shape matching rely on SHOT descriptors \cite{shot} as input signal for the neural network to produce correspondences between shapes \cite{halimi2019unsupervised, roufosse2019unsupervised, eisenberger2020deep}. This descriptor is orientation-aware but very sensitive to the input triangulation, resulting in overfitting to the training triangulation as demonstrated in \cite{donati2020deep}, and also in the first experiment of the main manuscript, with anisotropic remeshings.
    
    Therefore we use an input descriptor that is agnostic to the input triangulation so as to not overfit to it: the WKS descriptor \cite{aubry2011wave}, which is built using the eigenvectors and eigenvalues of the Laplace-Beltrami operator. These descriptor functions $(h_k)_{k \in [1,d]}$ are therefore fully intrinsic, and will display the same intrinsic self-symmetries as the shapes themselves. Namely, with the notations of Theorem \ref{thm:main}, $h_k \circ T = h_k$.
    
    Another commonly used option for an input signal is the 3-dimensional extrinsic coordinates of the shape points, as done in \cite{donati2020deep}. However, this input signal is dependent on the shape position in space. 
    With this input signal, the method is no longer fully intrinsic and therefore potentially unstable to rotations of the input shapes. Consequently, the input data needs to be centered, and augmented by adding randomly rotated poses.
    
    For this first ablation experiment, we train our method with this input signal (denoted as ``xyz input" in Table \ref{table:ablation_study}) instead of WKS descriptors. To make the experiment more complete, we report the results with two different data augmentations: \textit{a)} The general case, where there is no prior on the shapes alignment, so the data need to be augmented with all 3D rotations (3 parameters space). We denote this data augmentation as ``3 axis" in Table \ref{table:ablation_study}. \textit{b)} The special case where the input shapes are all aligned to one axis, but potentially rotated around this axis, so the data needs to be augmented around this axis (1 parameter space).  We denote this data augmentation as ``1 axis" in Table \ref{table:ablation_study}. We stress the fact that this kind of prior on the shapes rigid alignment already makes the method weakly supervised.
    
    We see in Table \ref{table:ablation_study} that even with the prior of shapes aligned to one axis, our method is better (and more general) when trained with WKS descriptors as an input signal.
    
\begin{figure}
\begin{center}
\includegraphics[width=1\linewidth]{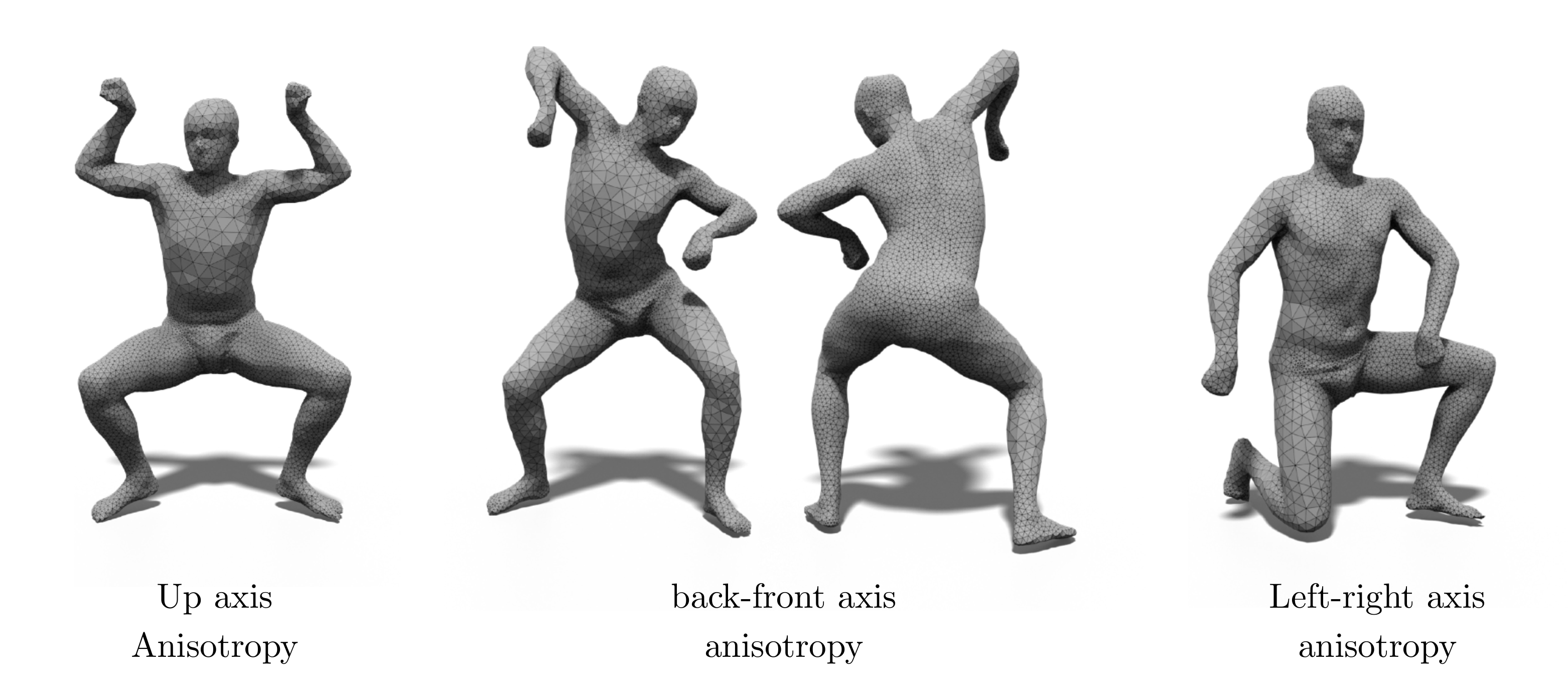}
\end{center}
\caption{SCAPE \cite{anguelov2005scape} dataset remeshed in an anisotropic fashion, used in the first experiment of the main manuscript
\label{fig:scape_aniso}}
\end{figure}
    
\subsection{The Orientation-aware Feature Extractor}
    To make our approach unsupervised, it is crucial that the feature extractor should be orientation-aware. Indeed, since we train on shapes exhibiting an isometric self symmetry (the left-right symmetry present in most organic shapes), the only way to disambiguate between left and right is through orientation, since the symmetric map reverses this orientation. DiffusionNet \cite{sharp2020diffusionnet} uses gradient features to incorporate this orientation information into potentially symmetric inputs (e.g. WKS descriptors in our case). For this second ablation, we propose to show that without this orientation-aware feature extractor, the method fails to produce informative descriptors, and report the results in Table \ref{table:ablation_study}, on row ``nonOA-FE" (standing for non orientation-aware feature extractor).
    
    To that end, we deactivate the gradient-based blocks of DiffusionNet, which results in a new orientation-agnostic feature extractor which can still produce excellent results \cite{sharp2020diffusionnet}. We then train our method using this feature extractor and WKS as input signal. We see in Table \ref{table:ablation_study} that this ablation greatly impairs the method.
    
    
\begin{figure}
\begin{center}
\includegraphics[width=1\linewidth]{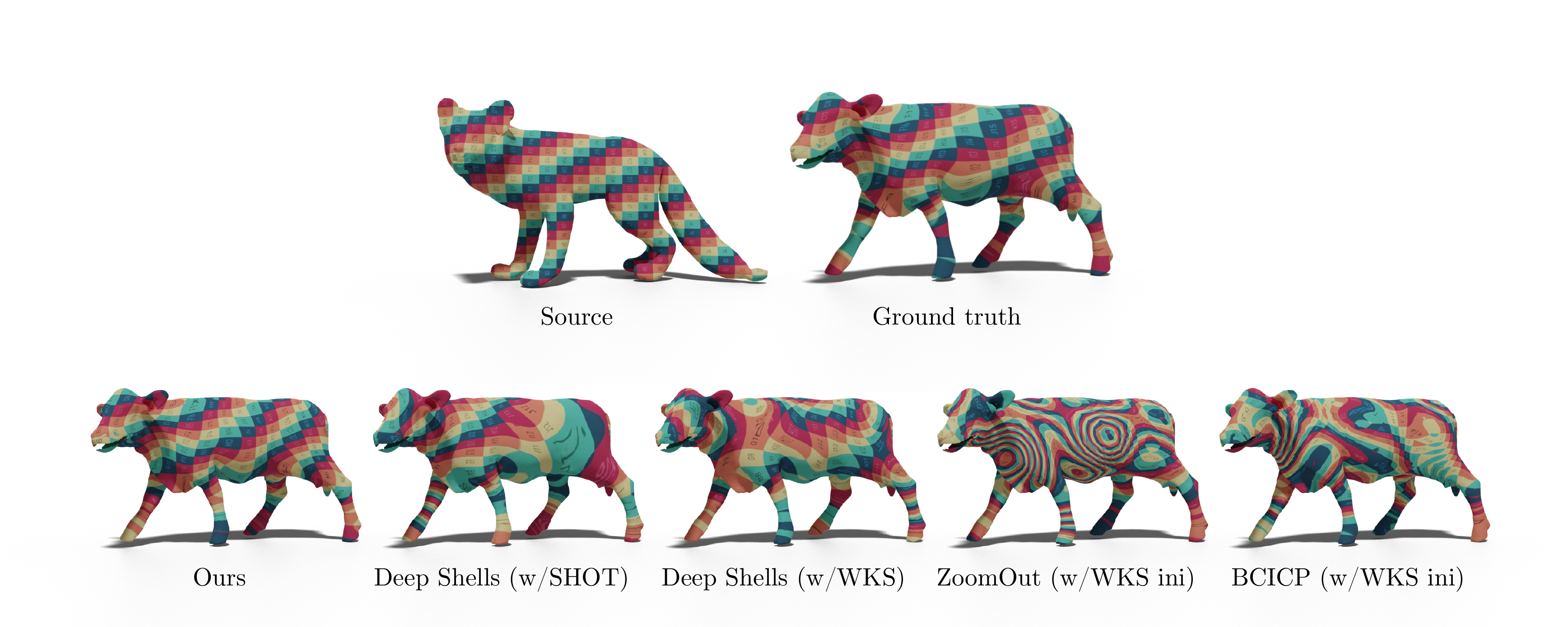}
\end{center}
\caption{Qualitative comparison to baselines on the SMAL dataset, using texture transfer from source to target shape. Only our method gives accurate correspondence, whereas in this challenging case baselines completely fail to predict the map.
\label{fig:smal_quali}}
\end{figure}

\subsection{The Complex Functional Maps Block and the Orientation-aware Loss}
    We remove the complex functional map block from the loss by setting $w_{\text{Q-ortho}} = 0$. As discussed in Theorem \ref{thm:main}, the resulting network is not guaranteed or encouraged to produce orientation-preserving correspondence. 
    We report the result of this ablation in Table \ref{table:ablation_study}, on rows ``no $Q$-maps".
    
    We observe that this ablation still seems to converge to well oriented maps in this case. This may be explained by the fact that DiffusionNet can produce non-symmetric descriptors from symmetric inputs like WKS, using shape orientation through gradients.
    Therefore, if two input shapes are consistently oriented, the symmetric input signal will be ``taken in the same direction" by DiffusionNet gradient-based blocks. Conversely, if two shapes are non-consistently oriented (e.g. one with inward normals, one with outwards normals), the symmetric input will be ``taken in opposite directions". In fact, using this remark one can retrieve symmetrized output descriptor functions (by symmetrized, here we mean composited with the intrinsic symmetric map $T_i$ of the shape $S_i$) generated by DiffusionNet from symmetric descriptors such as WKS, by simulating a change in shape orientation (which corresponds to a conjugation operation on the tangent bundle structure, or more practically to setting \texttt{gradY = -gradY} in DiffusionNet gradient operator entries).

    \begin{figure}
    \begin{center}
    \includegraphics[width=1\linewidth]{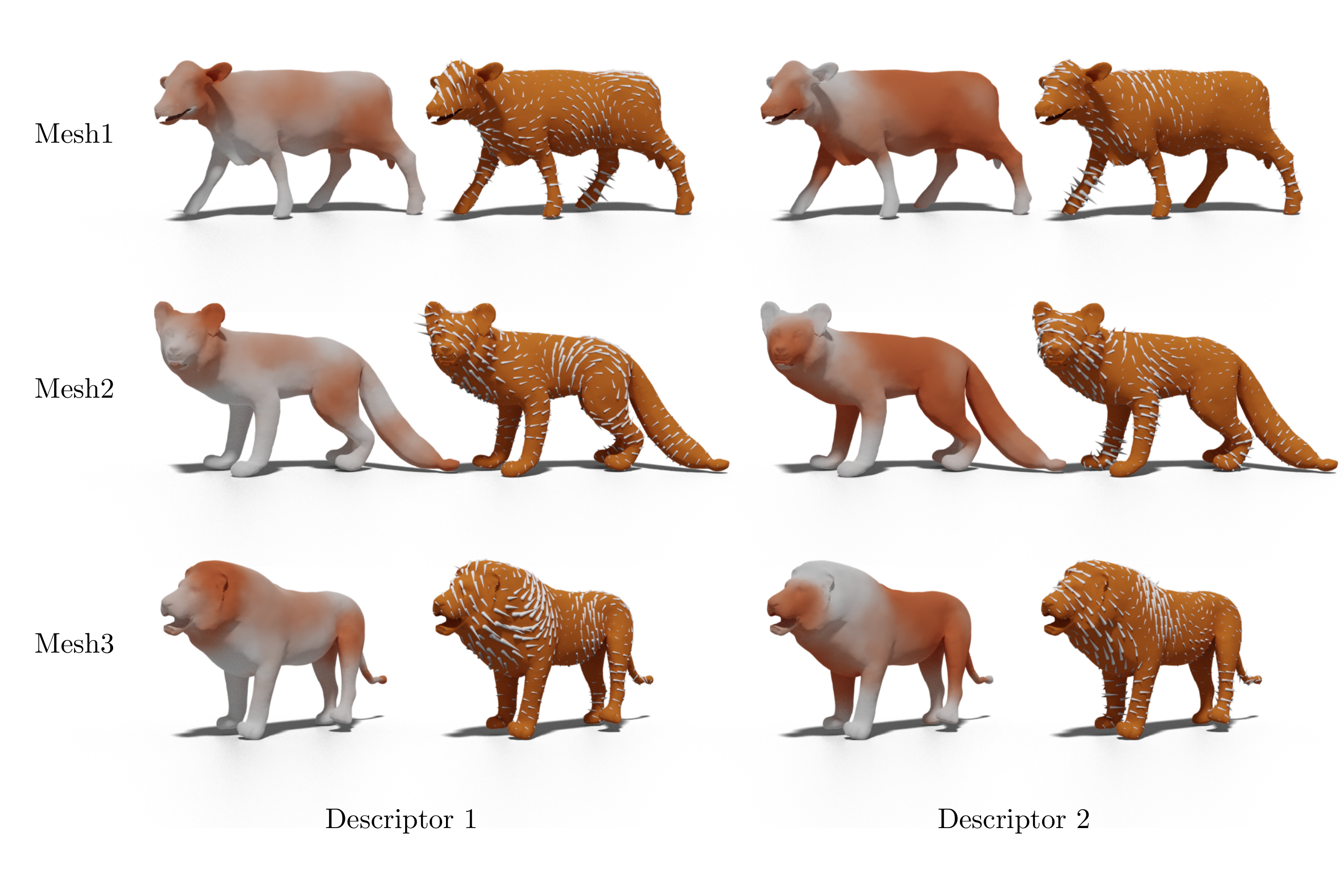}
    \end{center}
    \caption{Visualization of two different scalar descriptors learned by our method, along with their vector valued counterparts, on 3 meshes from the SMAL dataset. Contrary to the descriptors produced by \cite{donati2020deep}, these descriptors are fully intrinsic and generally not localised. However, we see that both our scalar and vector valued descriptors are robust to strong distortions.
    \label{fig:descs}}
    \end{figure}

    In practice, a second beneficial effect of our complex functional map loss is the reduction of overfitting.
    Indeed, in the experiment reported in Table \ref{table:smal} of the main manuscript, the train set is made of 32 SMAL re-meshed shapes and the test set is made of 17 shapes other SMAL re-meshed shapes. Learning methods are thus liable to overfit to their training set.
    We see in Table \ref{table:ablation_study} (where we report the geodesic error at epoch 3 and epoch 15 both with and without the complex functional map layer/loss) that without the complex functional map loss, the method is more prone to overfitting, as it looses generality if trained for too many epochs. To summarize, our complex functional map block and loss theoretically guarantee our approach to be orientation-preserving, and in practice also improve the pipeline stability with respect to overfitting.

\section{More Quantitative Results}

\begin{figure}
\begin{center}
\includegraphics[width=1\linewidth]{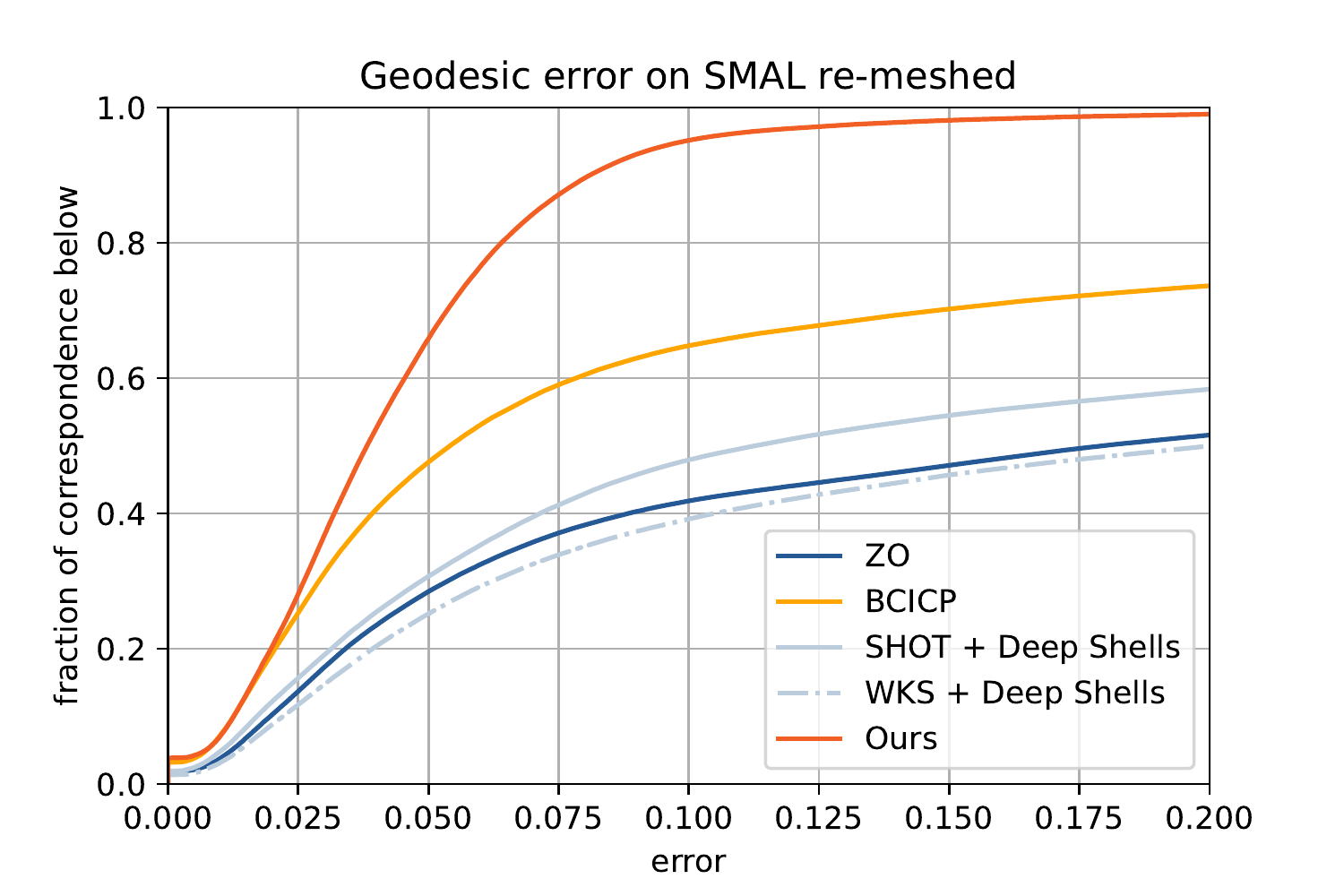}
\end{center}
\caption{Quantitative results of the different methods using the protocol introduced in \cite{kim11}, on the SMAL remeshed test set (third experiment of the main manuscript).
\label{fig:smal_quanti}}
\end{figure}

For completeness, we report in Figure \ref{fig:smal_quanti} the accuracy of our method and some baselines on the third experiment we conducted in the main manuscript (trained on 32 SMAL remeshed shapes, tested on 17 other SMAL remeshed shapes), using the evaluation protocol introduced in \cite{kim11}. We see that our method gives the best correspondence quality by far, as in this case it always predicts well-oriented maps for the test pairs (we see the tail of the error curve quickly reaches the $y=1$ line, which is equivalent to saying most predicted correspondences are extremely close to ground-truth).

\section{More Qualitative Results}

\subsection{Anisotropic Remeshing}

In Figure \ref{fig:scape_aniso}, we show the anisotropic remeshing of SCAPE dataset \cite{anguelov2005scape}, generated with Mmg \cite{dobrzynski2008anisotropic,dapogny2014three}. We use this anosotropic remeshing in the first experiment of the main manuscript to show that SHOT \cite{shot} based learning methods do not generalize to unseen triangulation.
Specifically, we see in Figure \ref{fig:scape_aniso} that the triangle scale is a function of the element coordinate.
For the first seven shapes of SCAPE test set, we constrain the triangle size to be dependent on the position on the up axis.
For the next seven shapes, we constraint the triangle size to be dependent on the position on the back-front axis.
For the remaining six shapes, we constraint the triangle size to be dependent on the position on the left-right axis.
With this remeshing, a network overfitting to the triangulation combinatorics will most likely fail to predict the desired map. Our method, which is triangulation agnostic, remains almost unaltered, as shown in the first experiment of the main manuscript.

\subsection{Another Qualitative Comparison on SMAL}

We report in Figure \ref{fig:smal_quali} a second texture transfer performed by baselines and our method on two of the SMAL test shapes. On this example the distortion between the two shapes is even stronger than on the example displayed in the main manuscript. However, our method still manages to predict accurate correspondences, while baselines fail to produce even a reasonable mapping in this case. Despite the fact that our method is spectral based, we see it can still produce accurate maps in challenging non-isometric cases.

\subsection{Visualization of the Scalar \& Vector Valued Descriptors Learned by our Method}

Lastly, we propose to visualize some descriptors learned by our network, also on the SMAL dataset, in Figure \ref{fig:descs}. Since our method also exploits the gradients of the scalar descriptors learned by DiffusionNet, we visualize these gradients (here rotated by $\pi/2$ to better make singularities stand out) which in fact correspond to the tangent vector field descriptors used to compute the complex functional map.
Our method enforces learned descriptors \emph{and their gradients} to correspond between source and target shapes, which was not done in any previous work to the best of our knowledge. Consequently, the features obtained with our method are all the more robust, since their gradients are also well-preserved under shape non-rigid deformation.

Indeed we notice that both the scalar-valued and the vector-valued part of the two descriptors displayed in Figure \ref{fig:descs} correspond well on the three chosen meshes, despite the strong distortions involved between these shapes.

\end{document}